\documentclass{article}
\usepackage{spconf,amsmath,graphicx}
\usepackage{amssymb}
\usepackage{changes}
\usepackage{comment}

\title{REAL-WORLD ATMOSPHERIC TURBULENCE CORRECTION VIA DOMAIN ADAPTATION}
\name{Xijun Wang$^1$, Santiago López-Tapia$^2$, Aggelos K. Katsaggelos$^2$\thanks{This work has been submitted for possible publication. Copyright may be transferred without notice, after which this version may no longer be accessible.}
\thanks{This research is based upon work supported in part by the Office of the Director of National Intelligence (ODNI), Intelligence Advanced Research Projects Activity (IARPA), via [2022-21102100007]. The views and conclusions contained herein are those of the authors and should not be interpreted as necessarily representing the official policies, either expressed or implied, of ODNI, IARPA, or the U.S. Government. The U.S. Government is authorized to reproduce and distribute reprints for governmental purposes notwithstanding any copyright annotation therein.}}
\address{$^1$Dept. of Computer Science, Northwestern University, Evanston, IL, USA\\
$^2$Dept. of Electrical and Computer Engineering, Northwestern University, Evanston, IL, USA}
\begin{document}
%
\maketitle
\begin{abstract}
Atmospheric turbulence, a common phenomenon in daily life, is primarily caused by the uneven heating of the Earth's surface. This phenomenon results in distorted and blurred acquired images or videos and can significantly impact downstream vision tasks, particularly those that rely on capturing clear, stable images or videos from outdoor environments, such as accurately detecting or recognizing objects. Therefore, people have proposed ways to simulate atmospheric turbulence and designed effective deep learning-based methods to remove the atmospheric turbulence effect. However, these synthesized turbulent images can not cover all the range of real-world turbulence effects. Though the models have achieved great performance for synthetic scenarios, there always exists a performance drop when applied to real-world cases. Moreover, reducing real-world turbulence is a more challenging task as there are no clean ground truth counterparts provided to the models during training. In this paper, we propose a real-world atmospheric turbulence mitigation model under a domain adaptation framework, which links the supervised simulated atmospheric turbulence correction with the unsupervised real-world atmospheric turbulence correction. We will show our proposed method enhances performance in real-world atmospheric turbulence scenarios, improving both image quality and downstream vision tasks.
\end{abstract}
\begin{figure}[htb]
\centering
\centerline{\includegraphics[width=8.5 cm]{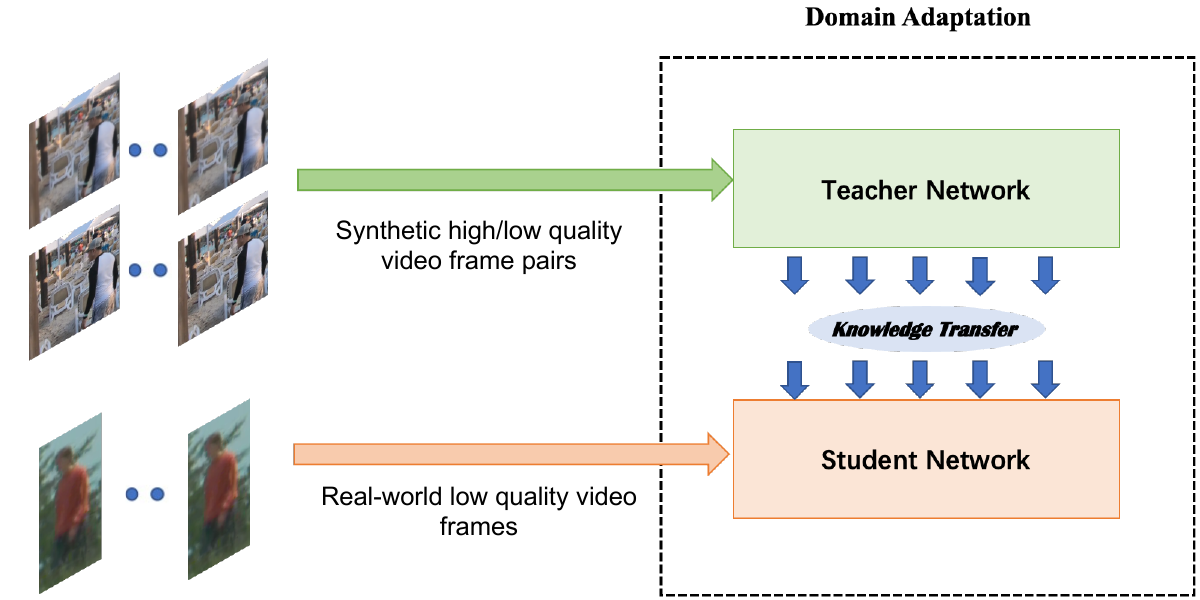}}
\caption{Domain adaptation framework.}
\label{fig:DA}
\end{figure}

\section{Introduction}
\label{sec:intro}
The effects of atmospheric turbulence (AT) can lead to several challenges, like image/video distortion and blurring, as well as reduced accuracy in subsequent vision tasks like object detection, recognition, or tracking. For instance, the task of identifying a person appearing in an image or video is important in many surveillance applications, and such surveillance tasks usually involve imaging over long distances, which are more susceptible to the effects of atmospheric turbulence \cite{nair2022comparison, chimitt2020simulating}. Therefore, multiple methods have been studied in the literature for AT mitigation. Initial efforts in AT correction followed adaptive optics-based approaches \cite{pearson1976atmospheric,roggemann2018imaging,tyson2022principles}, and these methods usually require expensive and complicated hardware.  Therefore, numerous image-processing-based methods were also developed \cite{aubailly2009automated, anantrasirichai2013atmospheric, zhu2012removing, chimitt2019rethinking}. These methods often involve combining the complementary clear regions across frames using the lucky fusion process, followed by deconvolution on the fused frame to obtain the restored output. \cite{nair2022comparison}. 

Recently, with the development of deep-learning (DL) algorithms for solving various inverse problems \cite{lucas2018using}, an increasing number of studies have proposed DL-based AT removal methods.  A deep learning-based approach is introduced in \cite{nair2021confidence} employing an effective nearest neighbors-based method for registration and an uncertainty-based network for restoration.  A physics-inspired transformer model for imaging through atmospheric turbulence is introduced in \cite{mao2022single}. The authors in \cite{zhang2024imaging} presented a multi-frame image restoration transformer tailored for addressing atmospheric turbulence. The authors in \cite{rai2022removing} employ a two-stage deep adversarial network aimed at minimizing atmospheric turbulence, where the first stage focuses on reducing geometrical distortion, while the second stage targets minimizing image blur.  A method leveraging well-trained Generative Adversarial Networks (GANs) as effective priors, learning to restore images within the semantic space context is proposed in \cite{mei2023ltt}. 
Reference \cite{lopez2023variational} proposes a variational deep-learning approach for video atmospheric turbulence mitigation. Finally, references \cite{wang2023atmospheric,nair2023ddpm,jaiswal2023physics} propose diffusion-based models for atmospheric turbulence restoration and generate high-quality images. All these current models are trained using synthetically generated data and tested on synthetic or real-world data. The performance of the existing models always decreases when applied to real-world images or videos. Therefore, inspired by the success of domain adaptation methods on real-world image super-resolution and blurring tasks \cite{liu2023real, wang2021unsupervised, wang2022semi}, to fill the gap between synthetic and real-world domains and effectively utilize the information from labeled synthetic data,  in this paper, we propose to use the domain adaptation technique to train the model with both synthetic and real-world data. As shown in Figure \ref{fig:DA}, we use a teacher-student framework and perform knowledge transfer from supervised learning to unsupervised learning. The teacher network is trained with a synthetic dataset containing simulated AT degraded frames paired with their corresponding ground-truth clean frames, enabling supervised learning. The student network is only trained with the AT degraded dataset from the real world, and no ground-truth clean frames are used, therefore the learning is unsupervised.

In summary, our main contributions are: 
\begin{itemize}
    \item We construct our AT correction model within a domain adaptation framework and bridge the gap between synthetic and real-world domains.
    \item We link the supervised simulated atmospheric turbulence correction with the unsupervised real-world atmospheric turbulence correction.
    \item Our proposed method helps achieve better performance in real-world atmospheric turbulence scenarios both in terms of image quality and the downstream vision task.
\end{itemize}

\section{Methodology}
\label{sec:methodology}
In Section \ref{ssec:DA-framework}, we  will first describe the design of the domain adaptation framework for our real-world AT mitigation model along with the associated objective functions. In Section \ref{ssec:Generator-framework}, we will delve into the design of our generators and their associated objective functions. Lastly, the summarized objective function during training, the training and testing procedures are outlined in Section \ref{ssec:train-test}.

\subsection{Domain adaptation framework}
\label{ssec:DA-framework}
\begin{figure*}[htb]
\centering
\centerline{\includegraphics[width=16.6 cm]{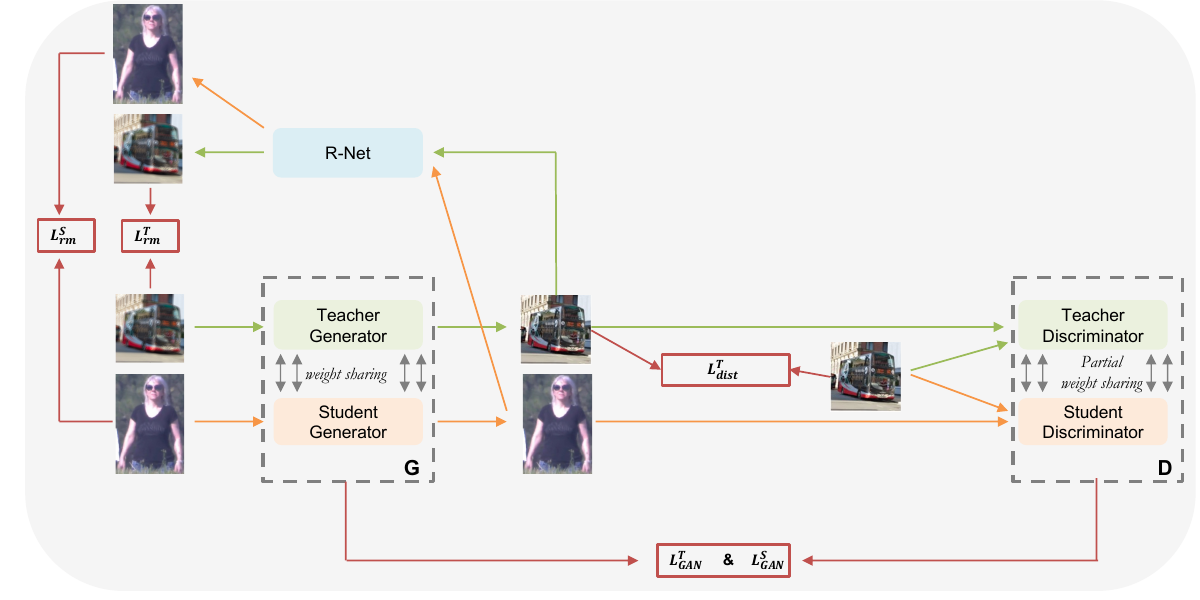}}
\caption{Real-world atmospheric turbulence mitigation model (Real-ATM) framework.}
\label{fig:realATM-framework}
\end{figure*}

The framework of our real-world atmospheric turbulence mitigation (Real-ATM) model is depicted in Figure \ref{fig:realATM-framework}. We employ a teacher-student (T-S) framework and facilitate knowledge transfer from supervised to unsupervised learning. Specifically, the teacher network is trained using a synthetic dataset comprising simulated AT-degraded frames and their corresponding ground-truth clean frames, thus learning in a supervised manner.  The student network is trained solely on the AT-degraded dataset from the real world, and there are no ground-truth clean frames for them. 

We adopt the GAN-based \cite{goodfellow2014generative} training to generate frames of high perceptual quality. First, for training the teacher component, we utilize pairs of synthetic AT-degraded and clean frames. The degraded frame \(y_T\) is fed into the teacher generator and outputs the restored one \(\hat{x}_T\). The teacher discriminator is trained to differentiate between its input being the restored frame \(\hat{x}_T\) generated by the teacher generator and the real frame \(x_T\) from the corresponding ground-truth clean frame. The Reproduce Net (R-Net) is trained to regenerate the degraded frame \(\hat{y}_T\)\added{,} taking the restored frame \(\hat{x}_T\) as input.

The objective function for training the teacher part first contains a pixel distance loss and a perceptual distance loss, measuring the distance between the restored frame and its corresponding ground-truth clean frame:
\begin{equation}
L^T_{dist} = \|x_T - G_{\theta}(y_T)\|_2^2 + \|\psi(x) - \psi(G_{\theta}(y_T))\|_2^2, 
\end{equation}
where \(x_T\) denotes the ground-truth clean frame, \(y_T\) represents the corresponding synthetic degraded frame, and \(G_{\theta}\) denotes the teacher generator, so \(G_{\theta}(y_T)\) represents the output of the generator, which is the restored frame \(\hat{x}_T\). The feature space denoted as $\psi(\cdot)$ is computed from the activations provided by the $3^{rd}$ and $4^{th}$ convolution layers of the $VGG$ network \cite{simonyan2014very}.  

Following \cite{goodfellow2014generative}, the adversarial losses for our AT mitigation task are shown as follows: The teacher generator network minimizes the following loss with respect to $\theta$
\begin{equation}
L^T_{gen} = \mathbb{E}_{y_T}\left[-\log
D_{\phi^T}\left(G_\theta\left(y_T\right)\right)\right],
\end{equation}
and the teacher discriminator network \(D_{\phi^T}\) minimizes the following loss with respect to $\phi^T$
\begin{equation}
\begin{split}
\small L^T_{dis} = & \mathbb{E}_{x_T}\left[-\log D_{\phi^T}\left(x_T\right)\right]+ \\
& \mathbb{E}_{y_T}\left[-\log\left(1-D_{\phi^T}\left(G_\theta\left(y_T\right)\right)\right)\right].
\end{split}
\end{equation}

A reproduce match loss is used to quantify the distance between the degraded frame reproduced by the R-Net and the true degraded frame:
\begin{equation}
L^T_{rm} = \|y_T - \hat{y}_T\|_2^2. 
\end{equation}

During the training of the student component, we utilize real-world AT data, which consists solely of degraded frames. During training, the student generator takes the real-world AT degraded frame \(y_S\)  as input and outputs the restored frame \(\hat{x}_S\). We have the student generator share weights with the teacher generator, constituting one of the knowledge transfer steps between the teacher and the student. 

The student discriminator \(D_{\phi^S}\) is trained to distinguish between the real clean frame \(x_S\) and the fake restored frame  \(\hat{x}_S\). Note that since we do not have the real clean frames, we utilize the clean frame \(x_T\)  from the dataset used for training the teacher part. This is feasible because the goal of the discriminator is to distinguish between data from two distributions: samples from the clean data distribution and those from the fake data distribution. Furthermore,  the student discriminator's weights are partially shared with the teacher discriminator's, representing the second stage of knowledge transfer. Finally, the restored frame \(\hat{x}_S\) is also fed into the R-Net, which regenerates the degraded frame \(\hat{y}_S\). The R-Net here serves as the third knowledge transfer stage.

The losses associated with the student part include the adversarial losses for the student generator and discriminator:
\begin{equation}
L^S_{gen} = \mathbb{E}_{y_S}\left[-\log
D_{\phi^S}\left(G_\theta\left(y_S\right)\right)\right],
\end{equation}
\begin{equation}
\begin{split}
L^S_{dis} = & \mathbb{E}_{x_T}\left[-\log D_{\phi^S}\left(x_T\right)\right]+ \\
& \mathbb{E}_{y_S}\left[-\log\left(1-D_{\phi^S}\left(G_\theta\left(y_S\right)\right)\right)\right],
\end{split}
\end{equation}
and the reproduce match loss between the regenerated degraded frame and the real degraded frame:
\begin{equation}
L^S_{rm} = \|y_S - \hat{y}_S\|_2^2. 
\end{equation}

Up to this point, we have gained a better understanding of how our model works via the knowledge transfer from supervised learning to unsupervised learning. Specifically, the T-S knowledge transfer relies on a shared generator network, partial weight sharing of dual discriminators, and a reproduce network. 

\subsection{Generator framework}
\label{ssec:Generator-framework}
\begin{figure}[htb]
\centering
\centerline{\includegraphics[width=6.0 cm]{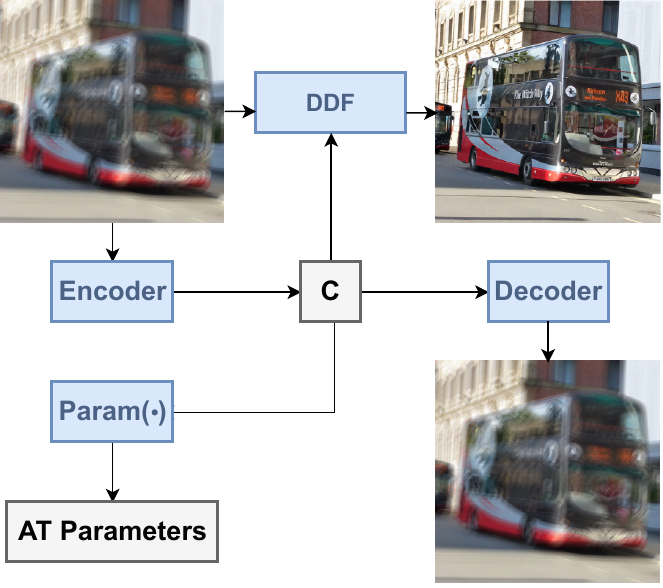}}
\caption{Generator training framework.}
\label{fig:generator-framework}
\end{figure}

Drawing inspiration from the variational inference framework employed in restoration models \cite{wang2023atmospheric, lopez2023variational, soh2022variational}, the detailed framework of the teacher and student generators is shown in Figure \ref{fig:generator-framework}. During training, the Decoupled Dynamic Filter (DDF)-based \cite{zhou2021decoupled} network aims at generating the restored frame from the degraded input frame conditioned on the latent feature \(c\). This \(c\) is learned via a variational autoencoder (VAE), which aims at learning to reconstruct the degraded frames, thereby imbuing \(c\) with image-prior information acquired from the degraded data. 

Additionally, we learn from the literature that incorporating the information from the degradation process during training could further improve the model’s performance. Therefore, we employ another CNN-based network to predict the atmospheric turbulence degradation parameters. Consequently, \(c\)  also encompasses prior information learned from the AT degradation procedure.

The losses associated with this VAE-based design contain a fidelity term and a reconstruct term. The fidelity term is the KL divergence of the learned posterior and the prior, which is a standard normal distribution. The reconstruction term measures the distance between the reconstructed degraded frame and the real degraded frame. 

Thus, for the teacher part, the VAE loss is defined as:
\begin{equation}
\label{eq:loss_vae_T}
L^T_{vae} = D_{KL}(q_{e_\psi}(c_T|y_T)||p(c_T)) + ||y_T - \hat{y}^G_T||_2^2, 
\end{equation}
where the first term is the fidelity term; it measures the fidelity of \(c_T\) extracted from the encoder \(e_\psi\), whose input is the degraded image \(y_T\). It is represented as the KL divergence of the approximate posterior \(q_{e_\psi}(c_T|y_T)\) from the prior \(p(c_T)\). We select the prior \(p(c_T)\) as a standard Gaussian distribution.  The second term is the reconstruction term, and we adopt the pixel-wise mean squared error (MSE) distance between the input degraded image \(y_T\) and the output \(\hat{y}^G_T\) of the decoder \(d_\varphi\). 

Similarly, the VAE loss for the student is:
\begin{equation}
\label{eq:loss_vae_S}
L^S_{vae} = D_{KL}(q_{e_\psi}(c_S|y_S)||p(c_S)) + ||y_S - \hat{y}^G_S||_2^2. 
\end{equation}

Finally, we would like the latent feature \(c\) to encompass knowledge about the AT degradation process. Remember that only the teacher knows the synthetic AT degradation process.  Therefore, we add a degradation loss for the teacher part, defined as:
\begin{equation}
\label{eq:loss_degrad}
L^T_{degrad} = ||\phi_{at} - \hat{\phi_{at}}||_2^2, 
\end{equation}
where \(\phi_{at}\) represents the ground-truth AT degradation parameters from the pre-trained AT simulator \cite{mao2021accelerating}. \(\hat{\phi_{at}} = Param(c_T)\) represents the estimated AT degradation parameters and is the output of a small network (parameter estimation module) \(Param(\cdot)\) taking \(c_T\) as input, as shown in Figure \ref{fig:generator-framework}.

\subsection{Training \& Testing}
\label{ssec:train-test}
Since we utilize a GAN-based training scheme, there is a generation procedure and a discrimination procedure, and they are trained alternatively \cite{goodfellow2014generative}. To summarize, the objective function during training contains both the teacher part and the student part. Specifically, during the generation procedure, the objective function is formulated as follows: 
\begin{equation}
\mathcal{L}_g = L^T + L^S,
\end{equation}
where 
\begin{equation}
\label{eq:L_T}
 L^T = L^T_{dist} + \lambda_1L^T_{gen} + \lambda_2L^T_{rm} + \lambda_3L^T_{vae} + \lambda_4L^T_{degrad},
\end{equation}
and 
\begin{equation}
\label{eq:L_S}
 L^S = \lambda_1L^S_{gen} + \lambda_5L^S_{rm} + \lambda_3L^S_{vae},
\end{equation}
where \(\lambda_1, \lambda_2, \lambda_3, \lambda_4\) and \(\lambda_5\) are hyper-parameters.
During the discrimination procedure,  the objective function is formulated as:
\begin{equation}
\mathcal{L}_d = L^T_{dis} + L^S_{dis}.
\end{equation}

\begin{figure}[htb]
\centering
\centerline{\includegraphics[width=6.0 cm]{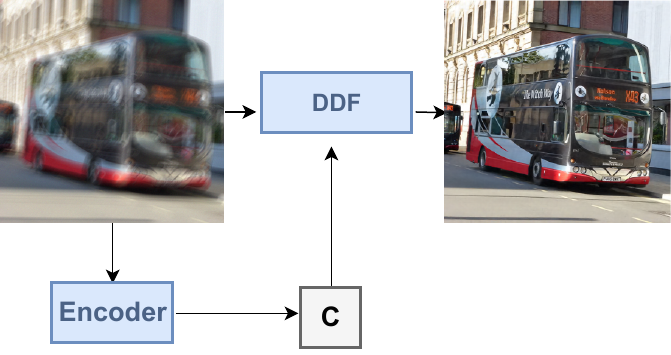}}
\caption{Testing framework.}
\label{fig:testing-framework}
\end{figure}

During testing,  only the generator network is needed. Illustrated in Figure \ref{fig:testing-framework}, the degraded frame is fed into the encoder module to obtain the latent feature \(c\). Then, \(c\) and the degraded frame are input into the DDF module to produce the restored output.

\section{Experiments}
\label{sec:experiments}
\subsection{Experimental settings}
\label{ssec:experimental-settings}
To get the synthetic training dataset for training the teacher part, we use the AT simulator in \cite{mao2021accelerating} to simulate the effect of AT on the REDS dataset \cite{nah2019ntire}, and the hyper-parameter of the simulator (\(D/r_0\)) is chosen randomly in the range [0.5, 2.0].  Our synthetic training dataset comprises $10^6$ AT-degraded and clean image pairs. We use another 2500 synthetic AT-degraded images as the synthetic testing dataset. For the student part training, we use 18 AT-degraded videos from the training set of the real-world dataset BRIAR \cite{cornett2023expanding}. We use another 54 AT-degraded videos from the testing set of BRIAR as the real-world testing dataset.

To construct our teacher and student generators, as shown in Figure 3, for the encoder module, we use 5 2D-convolution (2D-conv) layers with ReLU activation and one down-sampling layer after the first conv layer. For the decoder module, we use 5 2D-conv layers with ReLU activation and one up-sampling layer after the first conv layer. For the DDF-based module, we use 1 2D-conv layer followed by 8 DDF bottleneck blocks \cite{zhou2021decoupled} and 2 2D-conv layers. For our parameter estimation module, we simply use 2 2D-conv layers with LeakyReLU activation. 
\begin{figure*}[htb]
\centering
\centerline{\includegraphics[width=17.5 cm]{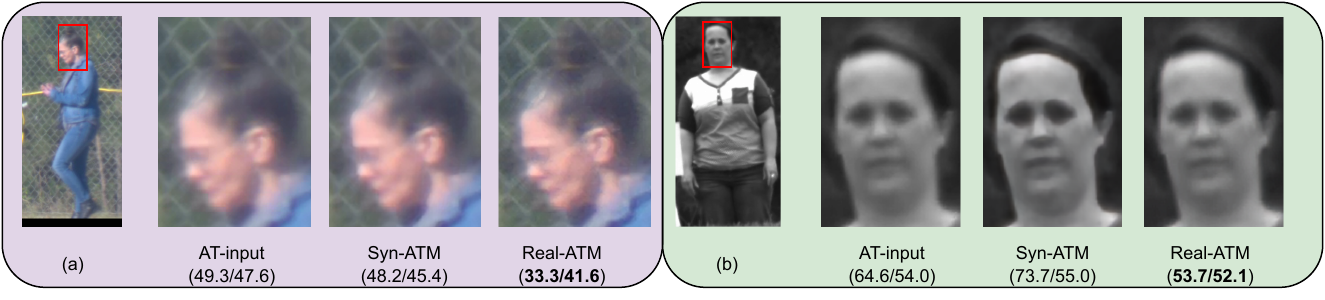}}
\caption{Visual samples of real-world AT-input, and the restored results from Syn-ATM and Real-ATM. (PIQE/BRISQUE) is shown in the parenthesis, the best score is marked in bold. }
\label{fig:visualsample}
\end{figure*}

To construct the teacher and student discriminators, we use  11 2D-conv layers with LeakyReLU activation and spectral normalization \cite{miyato2018spectral}. The partial weight sharing of the teacher and student discriminators occurs from the 6th to the 9th convolutional layers.

During training, we augment the training data by random cropping (\(160\times160\)), random vertical and horizontal flips, and random transposing. We train our model for 200 epochs with 1500 iterations per epoch, and we set the batch size to 16. We use the Adam optimizer \cite{kingma2014adam} with no weight decay, and we set the initial learning rate to 1\(e-4\) and gradually reduced it to 5\(e-6\) during training utilizing the cosine annealing schedule \cite{loshchilov2016sgdr}. The hyper-parameters \(\lambda_1, \lambda_2\),  \(\lambda_3\),  \(\lambda_4\),  and \(\lambda_5\) used in Equation \ref{eq:L_T} and Equation \ref{eq:L_S}  are set to 1e-3, 5e-1, 1e-1, 2e-1, and 2.5e-1, respectively. All the experiments are performed on two NVIDIA Quadro RTX 8000 GPUs.

\subsection{Results}
\label{ssec:results}
In this section, we will show that our proposed method enhances performance in real-world atmospheric turbulence (AT) scenarios, benefiting both image quality and the downstream person identification task.

\subsubsection{Image quality}
In Table \ref{tab:image-quality-syn}, we show the Peak Signal to Noise Ratio (PSNR) and the Structural Similarity Index Measure (SSIM) results of the model trained only with the AT synthetic dataset (Syn-ATM), i.e., only the teacher part is trained. We can see that the current model has achieved noticeable improvement on the synthetic testing set.  Due to the absence of ground-truth counterparts in the real-world testing set, we adopt the no-reference image quality metrics: Blind/Referenceless Image Spatial Quality Evaluator (BRISQUE) and Perception-based Image Quality Evaluator (PIQE) to evaluate the image quality (lower metric values indicate higher image quality). As shown in Table \ref{tab:image-quality-real}, for the Syn-ATM model, the image quality does not improve much when we test it on the real-world AT testing set. This observation aligns with our explanation that synthesized turbulent images are insufficient in capturing all real-world turbulence characteristics. 
\begin{table}[htbp]
\begin{center}
\small
\begin{tabular}{ccc}
 \hline
 & \textbf{PSNR\(\uparrow\)} & \textbf{SSIM\(\uparrow\)} \\ 
 \hline\hline
\textbf{AT-input} & 24.19 & 0.7263\\ 
 \hline
\textbf{Syn-ATM} & \textbf{30.61} & \textbf{0.8168}\\ 
 \hline
\end{tabular}
\end{center}
\vspace{-0.2 cm}
\caption{Image quality results of the Syn-ATM model on the synthetic AT testing set.}
\label{tab:image-quality-syn}
\end{table}
\begin{table}[htbp]
\begin{center}
\small
\begin{tabular}{ccc}
 \hline
 & \textbf{BRISQUE\(\downarrow\)} & \textbf{PIQE\(\downarrow\)} \\ 
 \hline\hline
\textbf{AT-input} & 47.67 & 63.39\\ 
 \hline
\textbf{Syn-ATM} & 46.99 & 68.89\\ 
 \hline
 \textbf{Real-ATM} & \textbf{44.90} & \textbf{49.95}\\ 
 \hline
\end{tabular}
\end{center}
\vspace{-0.2 cm}
\caption{Image quality results of Syn-ATM and Real-ATM models on the real-world AT testing set. The best result is marked in bold.}
\label{tab:image-quality-real}
\end{table}
Despite achieving remarkable performance in simulated scenarios, a performance drop occurs when applying the models to real-world cases. Conversely, our proposed real-ATM model obtains improved results on both metrics, demonstrating the effectiveness of our proposed method in filling up the domain gap between synthetic and real-world space.  Some real-world visual examples are shown in Figure \ref{fig:visualsample}. We could see that the real-ATM model provides better visual quality. Besides, we observed that the synthetic-ATM model can sometimes generate results with artificial characteristics and does not look like a naturally occurring image, as shown in Figure \ref{fig:visualsample} (b). This discrepancy arises because the model is solely trained on the synthetic AT dataset, failing to capture the full complexity of real-world AT distributions.

\subsubsection{Downstream person identification task}
According to previous works from the literature \cite{yang3rethinking, talebi2103learning, zhang2023darkvision}, image and video restoration as a fundamental low-level vision task can significantly improve the visual quality and benefit many downstream computer vision tasks. However, there is a complexity in real-world applications that might not be solved by quality alone. 
\begin{table}[htbp]
\begin{center}
\small
\begin{tabular}{ccc}
 \hline
 & \textbf{Top-20\(\uparrow\)} & \textbf{Top-15\(\uparrow\)} \\ 
 \hline\hline
\textbf{AT-input} & 18.5\% & 14.8\%\\ 
 \hline
\textbf{Syn-ATM} & 14.8\% & 7.4\%\\ 
 \hline
 \textbf{Real-ATM} & \textbf{25.9\%} & \textbf{16.7\%}\\ 
 \hline
\end{tabular}
\end{center}
\vspace{-0.2 cm}
\caption{Human recognition results on real-world AT videos and restored videos by Syn-ATM and Real-ATM. The best result is marked in bold.}
\label{tab:identification}
\end{table}
Sometimes the enhancement of the perceptual quality of images does not necessarily improve the performance on vision tasks or the enhancement strategies need to be tailored to specific applications rather than just improving image quality alone.  Therefore, in this section, we investigate the potential benefits of our proposed restoration network for downstream tasks. Specifically, we evaluate our model on a human recognition task to validate the necessity of restoration. For evaluation, we use the real-world person identification model, ShARc \cite{zhu2024sharc}, applied to our real-world testing set comprising 54 videos from the BRIAR testing set. As shown in Table \ref{tab:identification}, top-M represents whether the person could be correctly identified within the top M predictions made by the identification model. When applying the identification model to the restored videos generated from the Syn-ATM, the identification results do not improve. When applying the identification model to the videos restored from the real-ATM, the identification results significantly improve, demonstrating the effectiveness of our proposed method in benefiting the downstream vision task, in addition to improving the image quality alone. 
\section{Conclusion}
\label{sec:conclusion}
In this paper, we introduce a real-world atmospheric turbulence mitigation method. Our model employs a domain adaptation teacher-student framework, connecting supervised synthetic AT mitigation with unsupervised real-world AT mitigation. This approach enhances our ability to leverage information from labeled synthetic datasets while reducing the domain gap between synthetic and real-world data. Our final model, real-ATM, not only improves image quality in restoring real-world atmospheric turbulence data but also benefits downstream vision tasks. In the future, we will explore more possibilities for maintaining high image quality performance while tailoring the model to benefit more real-world applications. 

\bibliographystyle{IEEEbib}
\bibliography{strings,refs}

\begin{thebibliography}{10}

\bibitem{nair2022comparison}
Nithin~Gopalakrishnan Nair, Kangfu Mei, and Vishal~M Patel,
\newblock ``A comparison of different atmospheric turbulence simulation methods for image restoration,''
\newblock in {\em 2022 IEEE International Conference on Image Processing (ICIP)}. IEEE, 2022, pp. 3386--3390.

\bibitem{chimitt2020simulating}
Nicholas Chimitt and Stanley~H Chan,
\newblock ``Simulating anisoplanatic turbulence by sampling intermodal and spatially correlated zernike coefficients,''
\newblock {\em Optical Engineering}, vol. 59, no. 8, pp. 083101--083101, 2020.

\bibitem{pearson1976atmospheric}
James~E Pearson,
\newblock ``Atmospheric turbulence compensation using coherent optical adaptive techniques,''
\newblock {\em Applied optics}, vol. 15, no. 3, pp. 622--631, 1976.

\bibitem{roggemann2018imaging}
Michael~C Roggemann and Byron~M Welsh,
\newblock {\em Imaging through turbulence},
\newblock CRC press, 2018.

\bibitem{tyson2022principles}
Robert~K Tyson and Benjamin~West Frazier,
\newblock {\em Principles of adaptive optics},
\newblock CRC press, 2022.

\bibitem{aubailly2009automated}
Mathieu Aubailly, Mikhail~A Vorontsov, Gary~W Carhart, and Michael~T Valley,
\newblock ``Automated video enhancement from a stream of atmospherically-distorted images: the lucky-region fusion approach,''
\newblock in {\em Atmospheric Optics: Models, Measurements, and Target-in-the-Loop Propagation III}. SPIE, 2009, vol. 7463, pp. 104--113.

\bibitem{anantrasirichai2013atmospheric}
Nantheera Anantrasirichai, Alin Achim, Nick~G Kingsbury, and David~R Bull,
\newblock ``Atmospheric turbulence mitigation using complex wavelet-based fusion,''
\newblock {\em IEEE Transactions on Image Processing}, vol. 22, no. 6, pp. 2398--2408, 2013.

\bibitem{zhu2012removing}
Xiang Zhu and Peyman Milanfar,
\newblock ``Removing atmospheric turbulence via space-invariant deconvolution,''
\newblock {\em IEEE transactions on pattern analysis and machine intelligence}, vol. 35, no. 1, pp. 157--170, 2012.

\bibitem{chimitt2019rethinking}
Nicholas Chimitt, Zhiyuan Mao, Guanzhe Hong, and Stanley~H Chan,
\newblock ``Rethinking atmospheric turbulence mitigation,''
\newblock {\em arXiv preprint arXiv:1905.07498}, 2019.

\bibitem{lucas2018using}
Alice Lucas, Michael Iliadis, Rafael Molina, and Aggelos~K Katsaggelos,
\newblock ``Using deep neural networks for inverse problems in imaging: beyond analytical methods,''
\newblock {\em IEEE Signal Processing Magazine}, vol. 35, no. 1, pp. 20--36, 2018.

\bibitem{nair2021confidence}
Nithin~Gopalakrishnan Nair and Vishal~M Patel,
\newblock ``Confidence guided network for atmospheric turbulence mitigation,''
\newblock in {\em 2021 IEEE International Conference on Image Processing (ICIP)}. IEEE, 2021, pp. 1359--1363.

\bibitem{mao2022single}
Zhiyuan Mao, Ajay Jaiswal, Zhangyang Wang, and Stanley~H Chan,
\newblock ``Single frame atmospheric turbulence mitigation: A benchmark study and a new physics-inspired transformer model,''
\newblock in {\em European Conference on Computer Vision}. Springer, 2022, pp. 430--446.

\bibitem{zhang2024imaging}
Xingguang Zhang, Zhiyuan Mao, Nicholas Chimitt, and Stanley~H Chan,
\newblock ``Imaging through the atmosphere using turbulence mitigation transformer,''
\newblock {\em IEEE Transactions on Computational Imaging}, 2024.

\bibitem{rai2022removing}
Shyam~Nandan Rai and CV~Jawahar,
\newblock ``Removing atmospheric turbulence via deep adversarial learning,''
\newblock {\em IEEE Transactions on Image Processing}, vol. 31, pp. 2633--2646, 2022.

\bibitem{mei2023ltt}
Kangfu Mei and Vishal~M Patel,
\newblock ``Ltt-gan: Looking through turbulence by inverting gans,''
\newblock {\em IEEE Journal of Selected Topics in Signal Processing}, 2023.

\bibitem{lopez2023variational}
Santiago L{\'o}pez-Tapia, Xijun Wang, and Aggelos~K Katsaggelos,
\newblock ``Variational deep atmospheric turbulence correction for video,''
\newblock in {\em 2023 IEEE International Conference on Image Processing (ICIP)}. IEEE, 2023, pp. 3568--3572.

\bibitem{wang2023atmospheric}
Xijun Wang, Santiago L{\'o}pez-Tapia, and Aggelos~K Katsaggelos,
\newblock ``Atmospheric turbulence correction via variational deep diffusion,''
\newblock in {\em 2023 IEEE 6th International Conference on Multimedia Information Processing and Retrieval (MIPR)}. IEEE, 2023, pp. 1--4.

\bibitem{nair2023ddpm}
Nithin~Gopalakrishnan Nair, Kangfu Mei, and Vishal~M Patel,
\newblock ``At-ddpm: Restoring faces degraded by atmospheric turbulence using denoising diffusion probabilistic models,''
\newblock in {\em Proceedings of the IEEE/CVF Winter Conference on Applications of Computer Vision}, 2023, pp. 3434--3443.

\bibitem{jaiswal2023physics}
Ajay Jaiswal, Xingguang Zhang, Stanley~H Chan, and Zhangyang Wang,
\newblock ``Physics-driven turbulence image restoration with stochastic refinement,''
\newblock in {\em Proceedings of the IEEE/CVF International Conference on Computer Vision}, 2023, pp. 12170--12181.

\bibitem{liu2023real}
Hanzhou Liu, Binghan Li, Mi~Lu, and Yucheng Wu,
\newblock ``Real-world image deblurring via unsupervised domain adaptation,''
\newblock in {\em International Symposium on Visual Computing}. Springer, 2023, pp. 148--159.

\bibitem{wang2021unsupervised}
Wei Wang, Haochen Zhang, Zehuan Yuan, and Changhu Wang,
\newblock ``Unsupervised real-world super-resolution: A domain adaptation perspective,''
\newblock in {\em Proceedings of the IEEE/CVF International Conference on Computer Vision}, 2021, pp. 4318--4327.

\bibitem{wang2022semi}
Lin Wang and Kuk-Jin Yoon,
\newblock ``Semi-supervised student-teacher learning for single image super-resolution,''
\newblock {\em Pattern Recognition}, vol. 121, pp. 108206, 2022.

\bibitem{goodfellow2014generative}
Ian Goodfellow, Jean Pouget-Abadie, Mehdi Mirza, Bing Xu, David Warde-Farley, Sherjil Ozair, Aaron Courville, and Yoshua Bengio,
\newblock ``Generative adversarial nets,''
\newblock in {\em Advances in neural information processing systems}, 2014, pp. 2672--2680.

\bibitem{simonyan2014very}
Karen Simonyan and Andrew Zisserman,
\newblock ``Very deep convolutional networks for large-scale image recognition,''
\newblock {\em arXiv preprint arXiv:1409.1556}, 2014.

\bibitem{soh2022variational}
Jae~Woong Soh and Nam~Ik Cho,
\newblock ``Variational deep image restoration,''
\newblock {\em IEEE Transactions on Image Processing}, vol. 31, pp. 4363--4376, 2022.

\bibitem{zhou2021decoupled}
Jingkai Zhou, Varun Jampani, Zhixiong Pi, Qiong Liu, and Ming-Hsuan Yang,
\newblock ``Decoupled dynamic filter networks,''
\newblock in {\em Proceedings of the IEEE/CVF Conference on Computer Vision and Pattern Recognition}, 2021, pp. 6647--6656.

\bibitem{mao2021accelerating}
Zhiyuan Mao, Nicholas Chimitt, and Stanley~H Chan,
\newblock ``Accelerating atmospheric turbulence simulation via learned phase-to-space transform,''
\newblock in {\em Proceedings of the IEEE/CVF International Conference on Computer Vision}, 2021, pp. 14759--14768.

\bibitem{nah2019ntire}
Seungjun Nah, Sungyong Baik, Seokil Hong, Gyeongsik Moon, Sanghyun Son, Radu Timofte, and Kyoung Mu~Lee,
\newblock ``Ntire 2019 challenge on video deblurring and super-resolution: Dataset and study,''
\newblock in {\em Proceedings of the IEEE/CVF Conference on Computer Vision and Pattern Recognition Workshops}, 2019, pp. 0--0.

\bibitem{cornett2023expanding}
David Cornett, Joel Brogan, Nell Barber, Deniz Aykac, Seth Baird, Nicholas Burchfield, Carl Dukes, Andrew Duncan, Regina Ferrell, Jim Goddard, et~al.,
\newblock ``Expanding accurate person recognition to new altitudes and ranges: The briar dataset,''
\newblock in {\em Proceedings of the IEEE/CVF Winter Conference on Applications of Computer Vision}, 2023, pp. 593--602.

\bibitem{miyato2018spectral}
Takeru Miyato, Toshiki Kataoka, Masanori Koyama, and Yuichi Yoshida,
\newblock ``Spectral normalization for generative adversarial networks,''
\newblock {\em arXiv preprint arXiv:1802.05957}, 2018.

\bibitem{kingma2014adam}
Diederik~P Kingma and Jimmy Ba,
\newblock ``Adam: A method for stochastic optimization,''
\newblock {\em arXiv preprint arXiv:1412.6980}, 2014.

\bibitem{loshchilov2016sgdr}
Ilya Loshchilov and Frank Hutter,
\newblock ``Sgdr: Stochastic gradient descent with warm restarts,''
\newblock {\em arXiv preprint arXiv:1608.03983}, 2016.

\bibitem{yang3rethinking}
Huan Yang,
\newblock ``Rethinking image and video restoration: An industrial perspective,''
\newblock {\em restoration}, vol. 3, pp. 20.

\bibitem{talebi2103learning}
H~Talebi and P~Milanfar,
\newblock ``Learning to resize images for computer vision tasks. arxiv 2021,''
\newblock {\em arXiv preprint arXiv:2103.09950}.

\bibitem{zhang2023darkvision}
Bo~Zhang, Yuchen Guo, Runzhao Yang, Zhihong Zhang, Jiayi Xie, Jinli Suo, and Qionghai Dai,
\newblock ``Darkvision: A benchmark for low-light image/video perception,''
\newblock {\em arXiv preprint arXiv:2301.06269}, 2023.

\bibitem{zhu2024sharc}
Haidong Zhu, Wanrong Zheng, Zhaoheng Zheng, and Ram Nevatia,
\newblock ``Sharc: Shape and appearance recognition for person identification in-the-wild,''
\newblock in {\em Proceedings of the IEEE/CVF Winter Conference on Applications of Computer Vision}, 2024, pp. 6290--6300.

\end{thebibliography}

\end{document}